# Using Fitness Dependent Optimizer for Training Multi-layer Perceptron

*Dosti Kh. Abbas[1, a] Tarik A. Rashid[2,b], Karmand H. Abdalla[3,c], and Nebojsa Bacanin [4,d] , Abeer Alsadoon[5,6,7,8,e]*

[1] *Faculty of Engineering, Soran University, Kurdistan, Iraq*
[2] *Computer Science and Engineering Department, University of Kurdistan Hewler, Kurdistan, Iraq.*
[3] *Education and Psychology Department, College of Education, University of Raparin, Kurdistan, Iraq.*
[4] *Faculty of Informatics and Computing, Singidunum University, 11000 Belgrade, Serbia.*
[5] *School of Computing and Mathematics, Charles Sturt University, Sydney, Australia*
[6] *School of Computing Engineering and Mathematics, Western Sydney University, Sydney City Campus, Australia*
[7] *Information Technology Department, Asia Pacific International College (APIC), Sydney, Australia*
[8] *Kent Institute Australia, Information Technology Department, Sydney, Australia*

E-mail: [a] *dosti.abbas@soran.edu.iq* , [b] *tarik.ahmed@ukh.edu.krd*,
[c] *karmand.hussein@uor.edu.krd*, [d]*nbacanin@singidunum.ac.rs*, [e]*alsadoon.abeer@gmail.com*

## Abstract

This study presents a novel training algorithm depending upon the recently proposed Fitness Dependent Optimizer (FDO). The stability of this algorithm has been verified and performance-proofed in both the exploration and exploitation stages using some standard measurements. This influenced our target to gauge the performance of the algorithm in training multilayer perceptron neural networks (MLP). This study combines FDO with MLP (codename FDO-MLP) for optimizing weights and biases to predict outcomes of students. This study can improve the learning system in terms of the educational background of students besides increasing their achievements. The experimental results of this approach are affirmed by comparing with the Back-Propagation algorithm (BP) and some evolutionary models such as FDO with cascade MLP (FDO-CMLP), Grey Wolf Optimizer (GWO) combined with MLP (GWO-MLP), modified GWO combined with MLP (MGWO-MLP), GWO with cascade MLP (GWO-CMLP), and modified GWO with cascade MLP (MGWO-CMLP). The qualitative and quantitative results prove that the proposed approach using FDO as a trainer can outperform the other approaches using different trainers on the dataset in terms of convergence speed and local optima avoidance. The proposed FDO-MLP approach classifies with a rate of 0.97.

**Keywords:** Optimization, Multilayer perceptron, Fitness dependent optimizer, Training neural network, Academic student Performance.

## 1 Introduction

In the area of artificial intelligence, one of the most important developments is the neural network (NN). The fundamental theory of NNs was firstly proposed in 1943 [1]. The NNs imitate the neurons of the human brain to solve engineering classification problems for most of the parts. In the literature, various kinds of NNs were studied for instance Radial Basis Function network [2], Feedforward Neural Network (FFNN) [3], Recurrent Neural Network [4], Kohonen Self-Organizing Network [5], and Spiking Neural Networks [6].

Recently, feedforward neural networks, especially, multilayer perceptron neural networks have been broadly applied in various applications. In the past, the mainly utilized training approach is the gradient-based technique, which is defined as back-propagation. Therefore, in the use of this training algorithm, some inherent drawbacks are also regularly happened. In general, one point that has attracted many researchers recently and one of the complicated challenges for most of the parts in machine learning, which is considered is the learning process of artificial neural networks. The major problem of training NNs is considered as the nonlinear nature and undefined best weights and biases array. Training an NN is an optimization task as it is required to search for the optimal array of weights of an NN in the process of training. Slow convergence speed and local optima stagnation are the essential drawbacks of conventional training algorithms. Hence, researchers use evolutionary algorithms to train NNs to beat these difficulties [7] [8] [9] [10].

The animals' behaviors and natural phenomena have been studied by researchers to sympathize with the way they find a solution for their problems. Such as the technique of ants to find their route, or how fishes hunt their prey. Therefore, these types of algorithms, are called nature-inspired algorithms. In the 1960s, developments in metaheuristic algorithms began at the University of Michigan. A genetic algorithm was published by John Holland and his partners in 1960 then republished it in 1970 and 1983 [11]. Thereafter, several algorithms have been studied, for example, simulated annealing by Kirkpatrick, Gelatt, and Vecchi [12], Particle Swarm Optimization (PSO) by James Kennedy and Russel C.

Eberhart [13], which has been employed for many certifiable applications, and Artificial Bee Colony (ABC) algorithm by Karaboga and Akay [14]. Currently, there are stronger algorithms in terms of exploration and exploitation such as the Whale Optimization Algorithm (WOA) [15], Donkey and Smuggler Optimization [16], grey wolf optimizer [17], fitness dependent optimizer [18], and so on. Furthermore, some novel algorithms published in recent years, which have had a great impact on improving the related research area, such as Slime Mould Algorithm (SMA) [32], Monarch Butterfly Optimization (MBO), Moth Search Algorithm (MSA) [34], Hunger Games Search (HGS) [35], Runge Kutta Method (RUN) [36], and Harris hawks optimization (HHO) [37].

In this research work, the FDO algorithm, which was recently proposed by Abdullah and Ahmed [18] is combined with MLP for optimizing weights and biases of the NN to predict outcomes of students. The reason that motivated us to use FDO to train MLP is because of the fast convergence speed and stability in the capabilities of exploration and exploitation in finding optimal weight arrays. FDO gives fast convergence in global optimization of just coverage of the search range, consequently, it uses the fitness function to generate appropriate weights, which helps the algorithm in the exploration and development stage. Another unique characteristic of FDO is that it stores the previous search agent speed for possible reuse in future steps. It can be thought of as a PSO-based algorithm because it uses a similar mechanism to update the agent's location. In the proposed approach, firstly, the FDO starts to initialize weights and biases for the MLP, after that a training dataset is fed to the MLP and its weights and biases will be optimized by the FDO. Then, the MLP is tested for evaluation using a pre-defined testing dataset. The qualitative and quantitative results prove that the presented approach using FDO as a trainer can exceed the other approaches using different trainers on the dataset in terms of convergence speed and local optima avoidance.

The other sections of this study are organized as follows. The literature work is discussed in section two. A brief introduction of MLP is represented in section three. In section four, FDO details are described. The proposed method of FDO-MLP is discussed in section five. In section six, experimental results are discussed. Finally, we conclude the study in section seven.

## 2 Literature Work

In the past and recent years, many types of researches have been done related to the optimization of NN's weights using metaheuristic optimization algorithms. For example, Zhang and Jing-Ru [19] proposed a hybrid algorithm consisted of a PSO algorithm combined with the BP algorithm to train the FFNN weights that gain some success in convergent accuracy and convergent speed. This approach could make use of the powerful local searching ability of the back-propagation algorithm. The researchers adopt a heuristic technique to present a transition from a search of form particle swarm to a search of form gradient descending. The main advantage of this hybrid technique is getting higher training accuracy compared with the BP algorithm and the PSO algorithm by using less CPU time. We can also say that their algorithm is more stable as it has a more smooth mean recognition rate. The main problem of this approach is that it is not applied to solve more practical problems. Also, Socha, Krzysztof, and Christian [20] presented a hybrid approach and employed it to classification problems from the medical field. Their approach incorporates short runs of classical gradient approaches for example backpropagation. The researchers obtained some good results, but the approach cannot solve big problems, additional studies are needed to prove how their approaches accomplish more complicated problems. To enhance the global optimization capabilities of this approach it is recommended to add mechanisms that allow a more efficient search space exploration. Moreover, to develop a new combination between two algorithms for forecasting wind speed depending on wavelet packet decomposition, Meng, Ge, Yin, and Chen [21] proposed to use a crisscross optimization algorithm combined with artificial NN. Also, Artificial Bee Colony was used by some researchers, such as Karaboga, Dervis, Akay, and Ozturk [22] and Taheri, Hasanipanah, Golzar, and Abd Majid [23] to train and optimize the weights of NN. The authors of the study [23] aimed to present a combination of two models for forecasting blast-produced ground vibration. There, the forecasted ground vibration values by artificial NN and ABC combined with artificial NN algorithms were compared with various models. Moreover, the Whale optimization algorithm was used as an optimizer for NNs in some studies, for instance, Aljarah, Ibrahim, Faris, and Mirjalili [24] proposed a technique based on the WOA for training a multi-layer perceptron network having a single hidden layer. The advantage of this method is that the local optimal avoidance rate is high and the speed of convergence is fast. Also, compared to current MLP training approaches, WOA can be more efficient and competitive. Finally, WOA can reliably train ANN with small or large connection weights and biases. However, we can use WOA to train other types of artificial neural networks. The application of WOA-trained MLP to engineering classification problems is worth considering. The use of WOA-trained MLP solving functions to approximate the data set is also a valuable contribution. Also, Alameer, Abd Elaziz, Ewees, Ye, and Jianhua [25] presented a new approach to predict fluctuations of monthly gold prices. The approach utilized WOA as a trainer to learn MLP. Furthermore, A modified version of GWO was proposed for training a multi-hidden recurrent NN named M-RNNGWO [9]. This approach outperformed the original version of GWO in terms of learning NN. The main advantage of this method is that it is more stable in finding over-fitting problems and dealing with local minimum problems.

In the literature, many studies have been proposed using data mining and machine learning algorithms in the educational field. For instance, Rao, Swapna, and Kumar [26] used several machine learning algorithms for predicting students campus placements such as J48, Random Tree, Naïve Bayes, and Random Forest in weka tool and Neural Network algorithms, Recursive Partitioning, Multiple Linear Regression, Regression Tree, and binomial logistic regression in R studio. In their experiments, it is proven that the random tree algorithm was able to obtain the best accuracy among the other algorithms used in their study. Also, the regression tree and recursive partitioning performed better accuracy. Also, Okubo, Yamashita, Shimada, and Ogata [27] proposed an approach to predict students' final grades using a recurrent neural network from the stored log data in the educational systems. It was confirmed that a recurrent neural network is powerful too early forecasting of final grades. The research was proposed to explore diverse factors predicted to influence students' performance in higher education. Moreover, BP was used to train an MLP for predicting student outcomes by Karamouzis and Vrettos [28]. The authors used several experiments to execute training and testing. The average classification rate was 77% of training and 68% of testing phases. For modeling student's academic profiles, a hybrid approach was proposed by Arora and Saini [29] using fuzzy probabilistic combined with the neural network. Their proposed technique allowed the performance prediction of students depending on some of the qualitative observations of students. The experimental results of the research showed the approach outperformed traditional BP as well as some statistical models. The results show that the FPNN training time is shorter and the test results are close to expectations. It demonstrates the advantages of this method and increases the ability of the network to make predictions more accurately. The proposed network shows that the average classification accuracy is 98.56%. Authors can improve the accuracy of classification by using different ANN architectures.

## 3 Multilayer Perceptron Neural Network

MLP is an FFNN having only one hidden layer, where FFNN is a neural network having only one-directional connections among the neurons or nodes. In FFNN, the nodes are sorted in different aligned layers [2]. The first and last layers are identified as the input and output layers respectively, Whereas, the hidden layers are those layers, which are located between the input and output layers. An MLP design is illustrated in Fig. 1.

For calculating the output of MLP, first, the inputs weighted sums are calculated using equation 1.

$$S_j = \sum_{i=1}^{n}(W_{ij} I_i) + \beta_j \qquad (1)$$

Where input variable $i$ is assigned as $I_i$, $W_{ij}$ is identified as the weighted connection between $I_i$ and the hidden node $j$, and $\beta_j$ is defined as the bias of the $j^{th}$ hidden neuron.

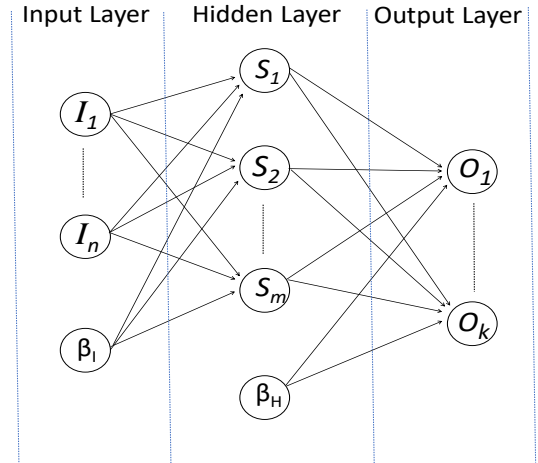

Fig. 1. Multilayer Perceptron Neural Network

The calculation of the output of the hidden nodes is shown in equation 2 by using an activation function. In this study, the sigmoid function is used.

$$f_j(x) = \frac{1}{1 + e^{-S_j}} \qquad (2)$$

The calculation of the final outputs is shown in equation 3. Its value is depending on the hidden neuron's calculated output. Where $W_{jk}$ indicates the weighted connection from the $j^{th}$ hidden neuron to the $k^{th}$ output neuron, and $\beta_k$ indicates a bias of the $k^{th}$ output neuron.

$$O_k = \sum_{i=1}^{m} W_{kj} f_i + \beta_k \qquad (3)$$

## 4 Fitness Dependent Optimizer Algorithm

FDO algorithm duplicates the process of reproduction of bees swarm. The method of scout bees searching is an important part of the FDO algorithm, which is taken from a new well-suited beehive among numerous possible beehives. In this algorithm, a possible solution is determined as scout bees searching for new beehives. Moreover, converging to the optimality can be considered by choosing the best beehive between several good beehives.

FDO gives fast convergence in global optimization of just coverage of the search range, consequently, it uses the fitness function to generate appropriate weights, which helps the algorithm in the exploration and development stage. Another unique characteristic of FDO is that it stores the previous search agent speed for possible reuse in future steps. It can be thought of as a PSO-based algorithm because it uses a similar mechanism to update the agent's location; however, FDO does it in a very different way. The statistics in the article show that FDO is superior to PSO, DA, GA, WOA, and SSA in many benchmarks. Generally, after testing many standard test functions and real applications, the number of search agents is found to have some relationship to the performance of FDO. Therefore, using a small number of agents (less than five) will significantly reduce the accuracy of the algorithm, while a large number of search agents will increase accuracy and spend more time and space.

FDO algorithm starts through randomly scout society initializing in the search area $X_i i(i = 1, 2, \ldots, n)$; the position of scout bees represents a newly found beehive, which is a solution. Finding better beehives is the aim of scout bees through randomly looking for more positions; the previously founded beehive is disregarded whenever a better beehive is found; therefore, each time the approach identifies an additional or better solution, that time the formerly found solution is disregarded. Nevertheless, in case of not being lead to a better solution in the former direction, it will then continue to the best solution that has been found, which is the current solution to that point.

In nature, searching for beehives by the scout bees is a random search. In the FDO algorithm, artificial scouts use the combination of fitness weight and a random walk mechanism for initially searching in the landscape. For that reason, the artificial scout bees change their position by inserting pace to the current position, scout bee desires to consider a better solution. Artificial scout bees' motion is represented as the equation (4). Where $x$ expresses a scout bee, which is defined as a search agent (scout bee), $i$ expresses the current search agent, $t$ is the current iteration, and $pace$; represents the direction of the scout bee and movement rate.

$$X_{i,t+1} = X_{i,t} + pace \qquad (4)$$

Mostly, $pace$ is based on the $fw$. Nevertheless, $pace$ direction is totally according to a random method. Therefore, for minimization problems, the $fw$ is calculated as equation (5). Where $x^*_{i,t\ fitness}$ represents the best global solution's fitness function value that has been found yet. $x_{i,t\ fitness}$ is defined as the value of the current solution's fitness function, and $wf$ is defined as a weight factor.

$$fw = \left|\frac{x^*_{i,t\ fitness}}{x_{i,t\ fitness}}\right| wf \qquad (5)$$

The weight factor value is determined as either zero or one to control the $fw$. It expresses a low chance of coverage and a high level of convergence when it is equal to 1. Nonetheless, if it is equal to 0, then it has not no effect on Equation (4), therefore it can be ignored. Moreover, it gives a more stable search. The value of fitness weight should be in the [0, 1] range; in whatever way, in some cases, fitness value is equal to 1, for instance, when the fitness value of the global best and the current solution is equal. Furthermore, a possibility is available that fitness value is equal to 0, which happens when $x^*_{i,t\ fitness}$ equals 0. Lastly, the division of any value by zero must be disregarded when $x_{i,t\ fitness}$ equals 0. Hence, rules 6 to 9 should be used. There, the value of $r$ is to be a random number in the range [-1, 1].

Regarding the mathematical complexity of the FDO algorithm: It has an $O\ (p * n + p * CF)$ time

$$\begin{cases} fw = 1\ or\ fw = 0\ or\ x_{i,t\ fitness} = 0, pace = x_{i,t} * r & (6) \\ fw > 0\ and\ fw < 1 \begin{cases} r < 0, pace = (x_{i,t} - x^*_{i,t}) * fw * -1 & (7) \\ r \geq 0, pace = (x_{i,t} - x^*_{i,t}) * fw & (8) \end{cases} \end{cases}$$

$$fw = \left|\frac{x_{i,t\ fitness}}{x^*_{i,t\ fitness}}\right| - wf \qquad (9)$$

complexity for each iteration, where $p$ denotes the population size, $n$ denotes the problem dimension, and the cost objective function ($CF$). Whereas, it has an $O\ (p * CF + p * pace)$ space complexity for all iterations, where the $pace$ identified as the best former paces stored. In this matter, the time complexity of the FDO algorithm is proportional to the iterations' number. Nevertheless, its space complexity has the same value during iterations [18].

## 5 Research Methodology

The proposed approach developed an MLP model trained by FDO for optimizing the weights and biases of the NN. In general, the problem is defined, then the data will be preprocessed, after that, we apply the process of feature selection. In the classification stage, NN is used as a classifier.

In the past, Rashid and Abbas [9], and Rashid and Aziz [30] have employed good research work on this dataset. It was collected from the University of Salahaddin - Erbil, which is comprised of 287 data samples and 21 features. The main features are consist of some personal information such as gender, age, and address. Also, some educational backgrounds and environmental causes are seen in features such as high school address, the language of education, and the educational level of parents. Moreover, some features are related to the institutional and achievements of the students such as department, quality assurance points of lecturer, overall score at the national exam, the course test score in a general university exam, and so on.

In this study, FDO-MLP is used to classify students' outcomes in a specific module. This process falls into two steps. At first, the MLP with one hidden layer will be trained through a training dataset. The NN's weights and biases will be optimized using the FDO. Secondly, the designed FDO-MLP is tested by a testing dataset for the evaluation of the trained model. In the design of this technique, two significant characteristics are contemplated. The first is the search agent representation in the FDO and the second one is the selection feature of the fitness function.

In this approach, to denote a candidate neural network, each scout bee will be encoded as a single-dimensional vector. This vector is divided into three sections, which are: weights for connecting the input and hidden layers, weights linking the hidden and output layers, and biases see Fig. 2.

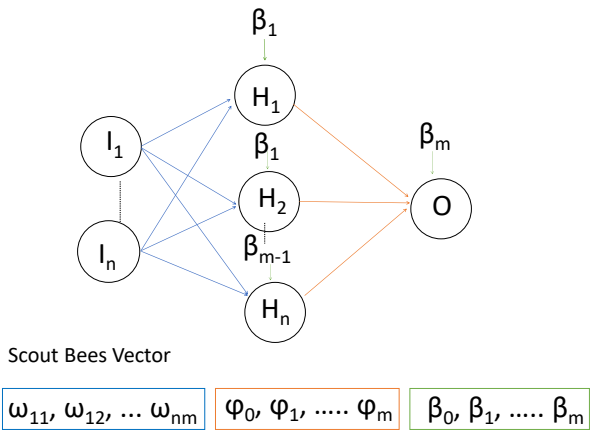

Fig. 2. Assigning an FDO scout bee vector to an MLP

The vector length equals the whole number of weights and biases in the NN. Equation (10) is used to calculate the length of the vector.

$$Vector\ Dimension = (n+1)*m + (m+1)*o \quad (10)$$

where $n$ is identified as the input nodes number, $m$ is indicated as the number of the hidden neurons, and $o$ is the number of output classes.

The fitness function Mean Square Error (MSE) is utilized for calculating the fitness value of search agents. The value of MSE is obtained in the difference between the actual value and predicted value through the produced agents (MLPs) for the whole training data set. Equation (11) is used to find the value of MSE

$$MSE = \frac{1}{n}\sum_{j=1}^{n}(O_j^k - d_j^k)^2 \quad (11)$$

Where $n$ is the data instance numbers in the training data set. When the $K^{th}$ training sample is calculated, $d_j^k$ denotes the predicted value of $j^{th}$ input neuron, and $O_j^k$ is defined as the actual value of $j^{th}$ input neuron.

The steps of the FDO-MLP technique are illustrated in Fig. 3, which are described in this paragraph. For training the MLP, FDO starts to randomly generate a predefined number of scout bees. Each scout bee demonstrates a possible MLP network. Here, the fitness function is employed to evaluate the MLP's quality. To perform this fitness evaluation, the produced search agent's vectors, which include weights and biases are first given to the MLP networks. After that, the networks are evaluated. The most used fitness function in evolutionary NNs is MSE, which we select in this work. The smallest MSE value is the target of the training algorithm in the MLP network depending upon dataset samples. Now, the scout bees restart to update their position. These processes are repeated, which are fitness evaluation and position updating of scout bees whenever the largest iterations number is achieved. In the end, the NN with the smallest value of MSE experiments on the predefined testing dataset.

In this approach, the student data has been normalized and the features were been suited to the main parameters of the FDO and the neural network. The number of connections between the neurons of the NN was indicated based on the number of features of the data. To tune the main parameters of FDO, the produced weight of the NN will be tested and given to the FDO. The proposed model has been tested with a different number of scout bees and iterations. In the results section, the values of these parameters were discussed and shown in Table 7.

# 6 Results and Discussion

In this research work, MATLAB R2014b has been used for all the experiments to perform the proposed trainer and other algorithms.

A suitable format for the data is provided using data pre-processing. After data pre-processing, the feature selection process is coming. The most related attributes to the class labels will be selected.

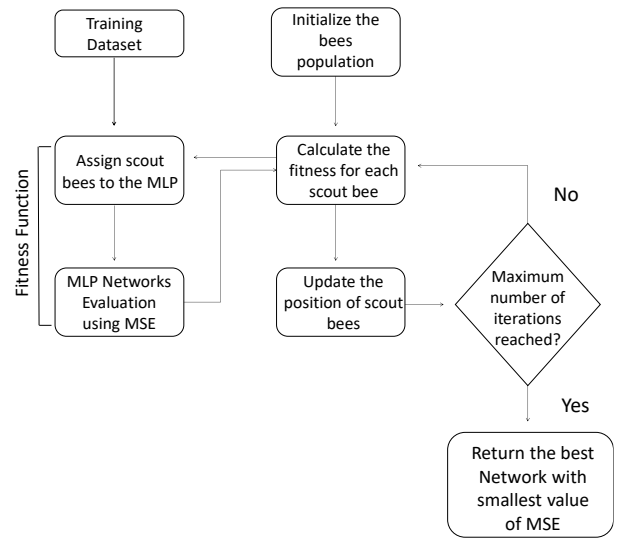

Fig. 3. Steps of the FDO-MLP approach

Determining the features makes the network have a higher accuracy rate because it eliminates the features with less related to the performance. In this study, those features that have not no impact on the output are removed. Moreover, only one attribute is chosen from those that have a similarity in affecting the output. The dataset was divided into 80% for training and 20% for testing. Five different runs including 200 iterations are executed for the experiments.

To choose several nodes in the hidden layer for MLP, different approaches were proposed. Nevertheless, there is no standard technique that is agreed upon about its dominance. In this study, several nodes will be selected in the hidden layer depending on this formula: $2 \times N + 1$ which is followed by the method utilized by Wdaa and Sttar [31], where $N$ is several features in the dataset.

Statistical results of the FDO-MLP approach are represented in Table 1. For each run time, the evaluation measures such as MSE, classification accuracy (AUC), sensitivity, specificity, and accuracy are shown for both training and testing phases.

The proposed FDO-MLP was compared with BP-MLP and some trainers depending on MSE and classification accuracy evaluation measures. Statistical results are demonstrated in Table 2, including the average (AVG), mean square error (STD), and the best and worst of classification accuracy. Also, it contains the most accurate result of FDO-MLP, BP, and FDO with cascade MLP (FDO-CMLP), GWO combined with MLP (GWO-MLP), modified GWO combined with MLP (MGWO-MLP), GWO with cascade MLP (GWO-CMLP), and modified GWO with cascade MLP (MGWO-CMLP). As demonstrated in the table, our proposed trainer FDO outperforms BP and all the other algorithms.

The best accuracy acquired by the FDO trainer confirms improvements among other algorithms employed. Since it's obtained low MSE and a high average of the classification proves that this technique can develop the best optimal array values for the MLP's weights and biases. Also, preventing convergence to the local optima.

Figure 4 shows the classification convergence curves using FDO-MLP, FDO-CMLP, GWO-MLP, MGWO-MLP, GWO-CMLP, and MGWO-CMLP depending on MSE averages for the training sample over five runs. The figure illustrates that FDO has a faster convergence speed than the other algorithms.

Table 1. Statistical results of the proposed FDO-MLP approach

| Run No. | Training/Testing | MSE | AUC | Sensitivity | Specificity | Accuracy |
|---|---|---|---|---|---|---|
| 1 | Training | 0.003578 | 0.95217 | 0.95302 | 0.95061 | 0.95217 |
|   | Testing | 0.00210 | 1.00000 | 1.00000 | 1.00000 | 1.00000 |
| 2 | Training | 0.0038306 | 0.97826 | 0.97315 | 0.98765 | 0.97826 |
|   | Testing | 0.00215 | 0.98245 | 0.97368 | 1.00000 | 0.98245 |
| 3 | Training | 0.003322 | 0.97391 | 0.97297 | 0.97561 | 0.97391 |
|   | Testing | 0.00246 | 0.96491 | 0.94871 | 1.00000 | 0.96491 |
| 4 | Training | 0.0034615 | 0.98695 | 0.98639 | 0.98795 | 0.98695 |
|   | Testing | 0.00202 | 1.00000 | 1.00000 | 1.00000 | 1.00000 |
| 5 | Training | 0.0042022 | 0.95217 | 0.94117 | 0.97402 | 0.95217 |
|   | Testing | 0.00234 | 0.91228 | 0.88095 | 1.00000 | 0.96815 |

Table 2. Statistical results of all the models

| Algorithm | FDO-MLP | BP-MLP | FDO-CMLP | GWO-MLP | MGWO-MLP | GWO-CMLP | MGWO-CMLP |
|---|---|---|---|---|---|---|---|
| MSE | 0.00221 | 0.00505 | 0.00268 | 0.00378 | 0.00354 | 0.00473 | 0.00501 |
| AVG | 0.97192 | 0.74736 | 0.92631 | 0.79649 | 0.85614 | 0.81052 | 0.76842 |
| Best | 1.00000 | 0.78947 | 0.98245 | 0.94736 | 0.87719 | 0.96491 | 0.68421 |
| Worst | 0.91228 | 0.70175 | 0.85964 | 0.64912 | 0.78947 | 0.73684 | 0.82456 |

Figure 5 shows the classification boxplots. In this study, the boxplots are used to demonstrate its information for 5 MSEs, which are provided by the optimizers at the end of the training stage. In the boxplot, the interquartile range relates to the box, farthest MSEs values are represented by the whiskers, the median value is represented by the bar in the box, and outliers are shown through the small circles. Here, the boxplots justify and prove the better performance of FDO to train MLP.

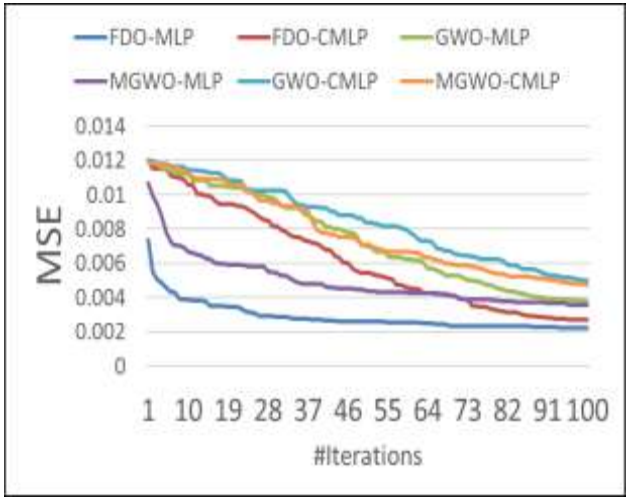

Fig. 4. MSE convergence curves

To validate the results and prove the performance of the proposed approach over the other approaches, we used cross-validation. Table 3 shows the results of classification by using cross-validation. The dataset was partitioned into five groups as follows X1, X2, X3, X4, and X5. The initial three groups comprised 57 data samples, and the last two groups contained 58 examples. In each case run, four groups were taken to the proposed system as the preparation dataset comprised of around 230 examples, what's more, the remaining were rolled, as the testing dataset comprised of around 57 examples to test the organization. The outcomes showed that the training classification rates in the folds were 97.82%, 96.95%, 96.52%, 92.57%, and 93.44%, and the average classification rate was 95.46%. Likewise, the classification rates for the testing stage for each crease were 96.49%, 94.73%, 96.49%, 91.37%, and 93.10%, and the average was 94.44%.

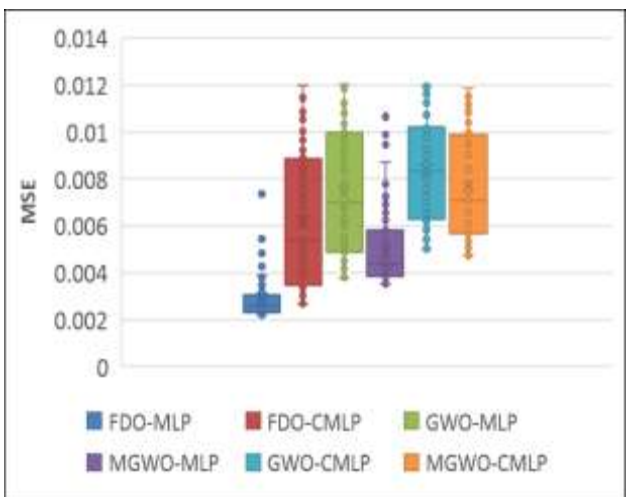

Fig. 5. MSE Boxplot charts

Table 4 demonstrates the outcomes of the students. It shows that there was a total of 287 students. The total number of students who failed is 104, the total number of students who passed the course is 183. In the first and third run, from the passing students, all the students were classified successfully out of 37 students, with a success rate of 100.00%, and only 2 students were not correctly classified out of 20 failed students with the success rate of 90.00%, In the second fold, 35 students were classified successfully out of 37 passed students. In contrast, only one student was not correctly classified in the group of failed students. In the fourth fold, 35 of 36 students who passed were classified correctly, resulting in a success rate of 97.22%. In contrast, 18 students out of 22 failing students were classified correctly as well. In the last turn, 34 and 20 students were classified successfully out of 36 and 22 passing and failing students respectively. In general, 178 students were classified successfully out of 183 passing students obtaining a 97.26% of success rate. On the other hand, 93 students were correctly classified out of 104 failing students with a success rate of 89.42%.

Figure 6 demonstrates that the FDO-MLP obtained the best accuracy among the other methods. The FDO-MLP was evaluated against some other techniques. The FDO-MLP was compared to FDO with Cascade MLP (FDO-CMLP). The FDO-MLP produced 94% accuracy, while the FDO-CMLP produced 90% accuracy. The modified GWO with Multilayer Perceptron (MGWO-MLP) obtained 87% accuracy, and the standard GWO with Cascade MLP (GWO-CMLP) obtained 86% accuracy. The GWO with MLP (GWO-MLP) produced 82% accuracy. However, the modified GWO with Cascade MLP (MGWO-CMLP) produced 72% accuracy.

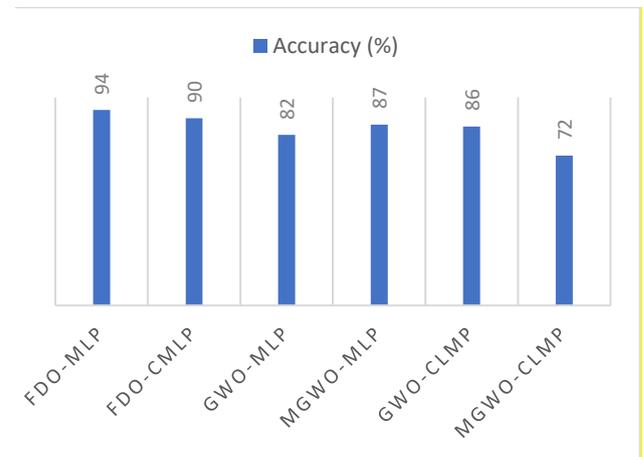

Fig. 6. Accuracy of the algorithms

In this research work, a confusion matrix was utilized as another measurement in the proposed approach to predict the classification results for the students. The testing results for the FDO-MLP will be assessed in the following paragraph.

Table 5 demonstrates the confusion matrix in the first fold for the FDO-MLP. The predicted number of true positives (passed) and the predicted number of false negatives (failed) were 37 and 2, respectively, the predicted number of false positives (passed) and the predicted number of true negatives (failed) were 0 and 18.

Table 3. Classification Results.

| Fold No. | Training/Testing | Dataset | No. Samples | MSE | Classification Rate |
|---|---|---|---|---|---|
| 1 | Training | X2+X3+X4+X5 | 230 | 0.0022817 | 97.82 % |
|   | Testing | X1 | 57 | 0.0033276 | 96.49 % |
| 2 | Training | X1+X3+X4+X5 | 230 | 0.0025636 | 96.95 % |
|   | Testing | X2 | 57 | 0.0032444 | 94.73 % |
| 3 | Training | X1+X2+X4+X5 | 230 | 0.0019557 | 96.52 % |
|   | Testing | X3 | 57 | 0.0033443 | 96.49 % |
| 4 | Training | X1+X2+X3+X5 | 229 | 0.002433 | 92.57 % |
|   | Testing | X4 | 58 | 0.0050468 | 91.37 % |
| 5 | Training | X1+X2+X3+X4 | 229 | 0.002523 | 93.44 % |
|   | Testing | X5 | 58 | 0.0042119 | 93.10 % |
| Average | Training |  |  | 0.002351 | 95.46 % |
|   | Testing |  |  | 0.003835 | 94.44 % |

Table 4. Performance and outcomes of the students.

| Fold No. | Data set | No. Samples | Passed Students ||| Failed Students |||
|---|---|---|---|---|---|---|---|---|
|  |  |  | No. Students | No. Correctly Classified Students | Success Rate | No. Students | No. Correctly Classified Students | Success Rate |
| Fold (1) | X1 | 57 | 37 | 37 | 100.00% | 20 | 18 | 90.00% |
| Fold (2) | X2 | 57 | 37 | 35 | 94.59% | 20 | 19 | 95.00% |
| Fold (3) | X3 | 57 | 37 | 37 | 100.00% | 20 | 18 | 90.00% |
| Fold (4) | X4 | 58 | 36 | 35 | 97.22% | 22 | 18 | 81.81% |
| Fold (5) | X5 | 58 | 36 | 34 | 94.44% | 22 | 20 | 90.90% |
| Total |  | 287 | 183 | 178 |  | 104 | 93 |  |
| Aver-age |  |  |  |  | 97.26% |  |  | 89.42% |

Table 6. Evaluation of the confusion matrix.

| Fold No. | Sensitivity | Specificity | PPV | NPV | Accuracy |
|---|---|---|---|---|---|
| 1 | 0.94 | 1.00 | 1.00 | 0.90 | 0.96 |
| 2 | 0.97 | 0.90 | 0.94 | 0.95 | 0.94 |
| 3 | 0.94 | 1.00 | 1.00 | 0.90 | 0.96 |
| 4 | 0.89 | 0.94 | 0.97 | 0.81 | 0.91 |
| 5 | 0.94 | 0.90 | 0.94 | 0.90 | 0.93 |
| Average | 0.93 | 0.94 | 0.97 | 0.89 | 0.94 |

Table 7. Weight complexity computation of the models.

| Algorithms | No. Connections | Search agents No. | No. Iterations | No. Hidden Layers | Testing Rate |
|---|---|---|---|---|---|
| FDO-MLP | 741 | 40 | 75 | 1 | 94.44% |
| FDO-CMLP | 760 | 40 | 75 | 1 | 90.57% |
| GWO-MLP | 741 | 50 | 75 | 1 | 82.52% |
| MGWO-MLP | 528 | 50 | 75 | 1 | 87.39% |
| GWO-CMLP | 544 | 50 | 75 | 1 | 86.42% |
| MGWO-CMLP | 544 | 50 | 75 | 1 | 71.79% |

Table 5. Confusion Matrix (Fold 1) for FDO-MLP

|  |  | Predicted | |
|---|---|---|---|
|  |  | Passed | Failed |
| Actual | Passed | 37 | 0 |
|  | Failed | 2 | 18 |

Notice from Table 6 that the value of Sensitivity, Specificity, Positive Predictive Value (PPV), Negative Predictive Value (NPV), and accuracy of the network was shown for all the folds using equations (12), (13), (14), (15), and (16) [9].

$$\text{Sensitivity} = \frac{TP}{TP + FN} \quad (12)$$

$$\text{Specificity} = \frac{TN}{TN + FP} \quad (13)$$

$$PPV = \frac{TP}{TP + FP} \quad (14)$$

$$NPV = \frac{TN}{TN + FN} \quad (15)$$

$$\text{Accuracy} = \frac{TP + TN}{TP + TN + FP + FN} \quad (16)$$

In the first turn, The sensitivity value was 0.94 indicating that the TPR was 94%, the specificity value was 1.00 indicating that the TNR was 100%, the PPV was 1.00 indicating that the success rate in passing students was 100%, the NPV was 0.9 indicating that the success rate in failing students was 90%, and the obtained accuracy of the network in the first fold was 96%.

Also, the results of the other folds were shown in table 6. The table contains the computation of the confusion matrix for FDO-MLP generally.

Table 7 demonstrates the problem dimension for the proposed approach compares to the other approaches. It is proven that the MLP outperforms the other neural network types if it is combined with FDO. Moreover, the FDO outperforms the other optimization algorithms if it is combined with all the types of neural networks. Whenever the FDO-MLP is used, the accuracy is greater than the one that uses other approaches. In the techniques, we used one hidden layer for all. The number of iterations is 75. The search agents' numbers are 40 for both FDO-MLP and FDO-CMLP and 50 for the other three approaches.

In general, it is proven that the FDO-MLP model can outperform other models in terms of convergence speed and avoiding convergence to local optima. The high exploration of the FDO algorithm produces high local optima avoidance. One of the significant reasons that helped the FDO algorithm in the training problem of multilayer perceptron to prevent numerous local solutions is by initializing artificial scout bees at random locations on the search space using upper and lower boundaries. Moreover, a special feature of the FDO algorithm is that if the current solution is better than the new solution, the artificial scout is continued by utilizing the former direction, only when it carries the scout to a better solution. FDO-based trainer originates a superior convergence speed from the saving of the best scout bee. Also, in case of not being lead the scout to a better solution by utilizing the former $pace$, then FDO preserves the current solution until the following cycle. Each time the solution is acknowledged, its pace esteem is put aside for potential reuse in the following cycle.

It is worth mentioning that the FDO approach additionally gives better outcomes using CMLP. Moreover, after FDO-MLP, the MGWO-MLP approach outperforms other models in terms of convergence speed as is shown in Figure 4.

# 7 Conclusion

In this study, the FDO algorithm is used to train MLP called the FDO-MLP approach. The main motivations to implement FDO to the training MLPs problem were the fast convergence speed and avoiding the convergence to local optima highly. Training MLP's problem was formed as the minimization problem. Getting to minimize MSE was the aim, and connection weights and biases were the parameters. The FDO algorithm was used to discover the optimal values for the weights and biases to minimize MSE. Moreover, the approach has been successfully applied to classify the outcomes of students with 0.97 classification accuracy.

In the results, it was shown that the proposed FDO-MLP approach can outperform other approaches in terms of also convergence, not only accuracy. The FDO provided superior results among BP and other evolutionary algorithms in terms of local optima avoidance and high exploration. Depending on the findings of this research work, we conclude that the FDO-MLP profits from a high local optima avoidance. Moreover, the convergence speed of the FDO trainer is high. Also, the FDO can be highly competitive and more efficient among the current training techniques of MLP. Finally, the FDO can train feedforward neural networks with a different number of weights and biases reliably. From the perspective of the mentioned points, we claim that the proposed approach can be effectively employed for predicting the outcomes of students.

For future work, this technique can be employed for other datasets in different fields. The applications of this approach are worth consideration in engineering classification problems. Also, it is recommended to train other kinds of artificial NNs using the FDO. It is worth mentioning that the FDO is a new, robust, malleable, and dynamic algorithm, which can easily be modified or hybridized with other algorithms. The FDO can be also hybridized with newer optimizers [38-41], such as Child

Drawing Development Optimization Algorithm Based on Child's Cognitive Development (CDDO), Formal context reduction in deriving concept hierarchies from corpora using adaptive evolutionary clustering algorithm star (ECA*), Dynamic Cat Swarm Optimization algorithm for backboard wiring problem (DCSO), and Learner Performance based Behavior algorithm (LBP).

## Biographies

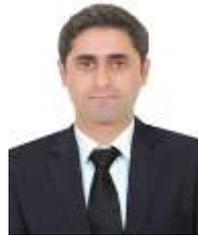

**Dosti Kh. Abbas** received his B.Sc. in Software Engineering degree from College of Engineering, Salahaddin University in 2008-2012. He pursued his Master's Degree at the Software Engineering, The Graduate School of Natural and Applied Sciences, Firat University, Turkey from 2015-2017. He started working as a Lecturer at Soran University in 2017.

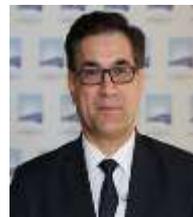

**Tarik A. Rashid** received his Ph.D. in Computer Science and Informatics degree from College of Engineering, Mathematical and Physical Sciences, University College Dublin (UCD) in 2001-2006. He pursued his Post-Doctoral Fellow at the Computer Science and Informatics School, College of Engineering, Mathematical and Physical Sciences, University College Dublin (UCD) from 2006-2007. Professor Rashid joined the University of Kurdistan Hewlêr (UKH) in 2017.

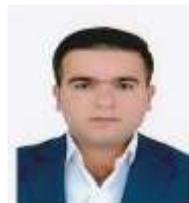

**Karmand H. Abdalla** received his B.Sc. in Computer Science degree from College of Education, Soran University in 2007-2011. He pursued his Master's Degree at the Software Engineering, The Graduate School of Natural and Applied Sciences, Firat University, Turkey from 2015-2017. He started working as a Lecturer at University of Raparin in 2017.

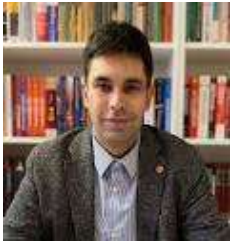
**Nebojsa Bacanin**, recived his PhD in Computer Science, Faculty of Mathematics, Univeristy of Belgrade. Professor Nebojsa does research in Algorithms, Databases and Artificial Intelligence. Their most recent publication is 'Bayesian methodology for target tracking using combined RSS and AoA measurements.

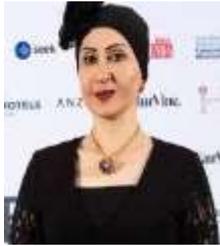
**Abeer Alsadoon** is an adjunct associate professor at Charles Sturt University (CSU) in Australia. She received her Ph.D. and Master Studies from the University of Technology, Baghdad in Iraq. Dr Alsadoon is academic experience includes working as an associate IT course coordinator and IT Senior lecturer at Charles Sturt University (CSU), Sydney (2012 – up to now), Researcher in Health Services Area, Researcher in e-health research group, and Researcher in Data Mining Area in CSU, Bathurst.